\def\year{2021}\relax
\documentclass[letterpaper]{article} 
\usepackage{aaai21}  
\usepackage{times}  
\usepackage{helvet} 
\usepackage{courier}  
\usepackage[hyphens]{url}  
\usepackage{graphicx} 
\usepackage{natbib}  
\usepackage{caption} 
\usepackage{amsmath}
\usepackage{amsfonts}
\usepackage{algorithm,algpseudocode}
\usepackage{threeparttable}
\usepackage{array}
\usepackage{subfig}
\usepackage{booktabs}

\title{Progressive Network Grafting for Few-Shot Knowledge Distillation}
\author {
    Chengchao Shen,\textsuperscript{\rm 1}
    Xinchao Wang, \textsuperscript{\rm 2}
    Youtan Yin, \textsuperscript{\rm 1}
    Jie Song, \textsuperscript{\rm 1}
    Sihui Luo, \textsuperscript{\rm 1} \\
    Mingli Song \textsuperscript{\rm 1, 3\thanks{Corresponding author}} \\
}
\affiliations {
    \textsuperscript{\rm 1} Zhejiang University,
    \textsuperscript{\rm 2} Stevens Institute of Technology, \\
    \textsuperscript{\rm 3} Alibaba-Zhejiang University Joint Research Institute of Frontier Technologies \\
    \{chengchaoshen, youtanyin, sjie, sihuiluo829, brooksong\}@zju.edu.cn, xinchao.wang@stevens.edu
}

\begin{document}

\maketitle

\begin{abstract}
Knowledge distillation has demonstrated encouraging performances in deep model compression. 
Most existing approaches, however, require massive labeled data to accomplish the knowledge transfer,  making the model compression a cumbersome and costly process.
In this paper, we investigate the practical \emph{few-shot} knowledge distillation scenario, 
where we assume only a few samples without human annotations 
are available for each category. 
To this end, we introduce a principled dual-stage distillation 
scheme tailored for few-shot data. In the first step, we
graft the student blocks one by one onto the teacher, 
and learn the parameters of the grafted block 
intertwined with those of the other teacher blocks. 
In the second step, the trained student blocks are progressively connected and then together grafted onto the teacher network, allowing the learned student blocks to adapt themselves to each other and eventually replace the teacher network. 
Experiments demonstrate that our approach, 
with only a few unlabeled samples,
achieves gratifying results on CIFAR10, CIFAR100, and ILSVRC-2012.
On CIFAR10 and CIFAR100, our performances are even
on par with those of knowledge distillation schemes
that utilize the full datasets.
The source code is available at \url{https://github.com/zju-vipa/NetGraft}. 
\end{abstract}

\section{Introduction}
{
    Deep neural networks have been widely applied 
    to various computer vision tasks, such as image
    classification~\cite{krizhevsky2012imagenet,simonyan2014very,szegedy2015going,he2016deep}, 
    semantic segmentation~\cite{long2015fully,chen2018deeplab,badrinarayananK17}, 
    and object detection~\cite{ren2015faster,liu2016ssd,zhang2018single}.
    The state-of-the-art deep models, however, 
    are often cumbersome in size
    and time-consuming to train and run, 
    which precludes them from being deployed in
    the resource-critical scenarios such as Internet of Things~(IoT).

    To this end, many model compression methods, such as network pruning~\cite{li2016pruning,han2015learning,wen2016learning} 
    and knowledge distillation~\cite{hinton2015distilling,romero2015fitnets}, 
    are proposed to trade off between model performance and size.
    Network pruning methods remove the weights of trained network
    based on  priors and then retrain the pruned network to recover
    the performance of the original network. 
    They require massive labeled data and an
    iterative retraining procedure, which is often time-consuming.
    Knowledge distillation methods~\cite{hinton2015distilling},
    on the other hand, {train student networks
    by making them imitate} the output of a given teacher.
    However, as student network is in many cases initialized randomly 
    and trained from scratch, 
    knowledge distillation approaches also rely on
    numerous training data to explore 
    the large parameter space of student so as to train a well-behaved model.

    To alleviate data hunger in knowledge distillation, 
    several few-shot distillation methods have been proposed 
    to transfer knowledge from teacher to student 
    with less dependency on the amount of data.
    The work of~\cite{li2018few} proposes a few-shot approach by 
    combining network pruning and block-wise 
    distillation to compress the teacher model.
    The one of~\cite{wang2020neural} introduces 
    an active mixup augmentation strategy that
    selects hard samples from a pool of the augmented images. 
    The work of~\cite{bai2019few} designs a 
    cross distillation method, combined 
    with network pruning to reduce the layer-wise
    accumulated errors in the few-shot setting.
    In spite of the encouraging results achieved,
    existing methods still heavily rely 
    on pre- or post-processing techniques,
    such as network pruning,
    which {\emph{per se}} are unstable and error-prone.
    
    In this paper, we propose a principled dual-stage 
    progressive network grafting strategy 
    for few-shot knowledge distillation,
    which allows us to eliminate the dependency on
    other techniques that are potentially fragile
    and hence strengthen the robustness of knowledge distillation.
    At the heart of our proposed approach is a
    block-wise ``grafting'' scheme, which 
    learns the parameters of the student network
    by injecting them into the teacher network
    and optimizing them intertwined with the 
    parameters of the teacher in a progressive fashion.
    Such a grafting strategy takes much better advantage of 
    the well-trained parameters of the teacher network
    and therefore significantly shrinks parameter space of the student network, 
    allowing us to training the student with much fewer samples.
    
    Specifically, our grafting-based distillation scheme follows
    a two-step procedure. In the first step, the student network is decomposed
    into several blocks, each of which contains fewer parameters to be optimized.
    We then take the block of the student to replace the corresponding
    one of the teacher, and learn the parameters of the student intertwined 
    with the well-trained parameters of the teacher, enabling the 
    knowledge transfer from the teacher to the student. 
    In the second step, the trained student blocks 
    are progressively connected and then grafted onto teacher network, 
    in which the student blocks learn to adapt each other and 
    together replace more teacher blocks. 
    Once all the student blocks are grafted onto the teacher network,
    the parameter learning is accomplished. 
    The proposed dual-stage distillation, by explicitly exploiting
    the pre-trained parameters and refined knowledge of the teacher,
    largely eases the student training process and reduces the 
    the risk of overfitting. 
    
    In sum, our contribution is a novel grafting 
    strategy for few-shot knowledge distillation, 
    which removes the dependency on other brittle 
    techniques and therefore reinforces robustness. 
    By following a principled two-step procedure, 
    the proposed grafting strategy dives into 
    the off-the-shelf teacher network
    and utilizes the well-trained parameters of the teacher
    to reduce the parameter search space of the student,
    thus enabling the efficient training the student under
    the few-shot setup. 
    With only a few unlabeled samples for each class, 
    the proposed approach achieves truly encouraging performances
    on CIFAR10, CIFAR100, and ILSVRC-2012.
    On  CIFAR10 and CIFAR100, the proposed method even yields results
    on par with those obtained by knowledge distillation using the full datasets.
    
}

\section{Related Work}
{
    {\bf Few-Shot Learning. } 
    The mainstream of few-shot learning research focuses on image classification, which learns to classify using few samples per category. 
    Two kinds of approaches are widely adopted: metric learning based method~\cite{koch2015siamese,vinyals2016matching,NIPS2017_6996} and meta learning based one~\cite{ravi2017optimization,wang2016learning,santoro2016one}. 
    The difference of problem setting between few-shot distillation and few-shot classification can be summarized as two fold.
    Firstly, a trained teacher model is available in few-shot distillation, but few-shot classification has none.
    Secondly, the model is trained on various related tasks in few-shot classification, but few-shot distillation is trained on the same task.

    \noindent{\bf Knowledge Distillation. } 
    Resuing pre-trained networks has recently attracted attentions from researchers in the field~\cite{Chen_2020_CVPR,Yu_2017_CVPR}.
    Bucilua et al~\cite{bucilua2006model} propose prototype knowledge distillation method, which trains a neural network using 
    predictions from an ensemble of heterogeneous models. 
    Hinton et al~\cite{hinton2015distilling} propose the knowledge distillation concept, where temperature is introduced to soften the predictions of teacher network. 
    Following~\cite{hinton2015distilling}, researchers pay more attention to the supervision from intermediate representations for better optimization performance~\cite{romero2015fitnets,Wang2018ProgressiveBK,shen2019amalgamating,shen2019customizing,ye2020data,luo2020collaboration}. 
    To reduce total training time, online distillation~\cite{yang2019snapshot,zhang2019your} is proposed to unify the training of student and teacher into one step. 
    The above methods consume tremendous labeled data to transfer 
    knowledge from teacher network, which significantly affects 
    the convenience of deployment in practice.
    The works of~\cite{Song_2019_NeurIPS,Song_2020_CVPR},
    on the other hand, focus on 
    estimating knowledge transferability 
    across different tasks, while
    those of~\cite{Yang_2020_CVPR,Yang_2020_NIPS}
    explore distillation on the graph domain.

    To reduce the data dependency, data-efficient and data-free knowledge distillation are investigated.
    FSKD~\cite{li2018few} adopts block-wise distillation to align the pruned student and teacher, where the blocks are not optimized to imitate the final teacher predictions.
    Wang et al~\cite{wang2020neural} combine image mixup and active learning to augment the dataset, which still requires considerable data. 
    ZSKD~\cite{nayak2019zero} adapts synthetic data impressions from teacher model to replace original training data to achieve knowledge transfer.
    ZSKT~\cite{micaelli2019zero} and DFAD~\cite{fang2019data} introduce an adversarial strategy to synthesize training samples for knowledge distillation. 
    DAFL~\cite{chen2019data} and KEGNET~\cite{yoo2019knowledge} are proposed to synthesize images by random given labels and then student learns to imitate teacher.

    However, data-free methods need to learn knowledge from massive low-quality synthetic samples, which is a time-consuming procedure.
    In the meanwhile, data-free methods need to redesign and train a dedicated generator to synthesize massive data, where image generation is largely limited by the capacity of the generator, especially for high-resolution images.
    Few-shot methods have higher training efficiency, which only involve few real samples during optimization, even only one sample.

    \noindent{\bf Grafting. }
    Grafting is adopted as an opposite operation against pruning in decision tree~\cite{webb1997decision,penttinen2003improving}, which adds new branches to the existing tree to increase the predictive performance. 
    Li et al~\cite{li2012network} graft additional nodes onto the hidden layer of trained neural network for domain adaption.
    NGA~\cite{hu2020exploiting} is proposed to graft front end network onto a trained network to replace its counterpart, which adapts the model to another domain. 
    Meng et al~\cite{meng2020filter} propose an adaptive weighting strategy, where filters from two networks are weighted and summed to reactivate invalid filters. 
    Our proposed network grafting strategy replaces cumbersome teacher blocks with the corresponding lightweight student ones in a progressive manner, which aims to smoothly transfer knowledge from teacher to student.
}

\section{The Proposed Method}
{
    \subsection{Overview}
    {
        \begin{figure*}[t]
        \centering
        \includegraphics[width=0.9\linewidth]{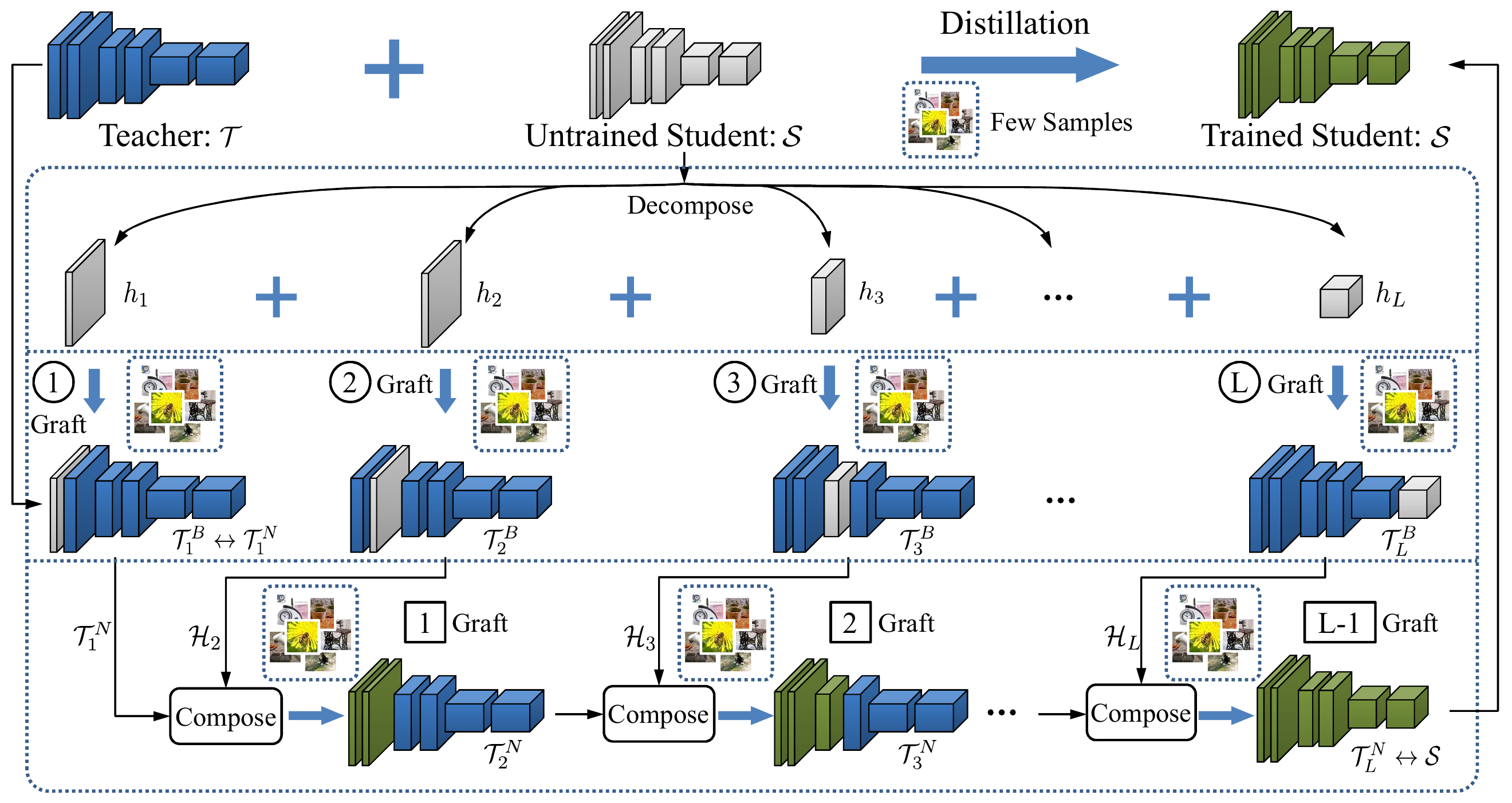}
        \caption{
        The dual-stage knowledge distillation strategy for few-shot knowledge distillation.
        Firstly, student network $\mathcal{S}$ is decomposed into several blocks: $\{h_l\}_{l=1}^L$, each of which is grafted onto teacher and then optimized by few-shot distillation between $\mathcal{T}_l^{B}$ and $\mathcal{T}$.
        Secondly, trained blocks in the first stage: $\{\mathcal{H}_l\}_{l=1}^L$ are sequentially composed into the trained student network: $\mathcal{S}$ using few-shot distillation between $\mathcal{T}_l^{N}$ and $\mathcal{T}$.
        }
        \label{fig:overview}
        \end{figure*}

        The goal of few-shot knowledge distillation is to transfer knowledge from teacher network $\mathcal{T}$ to student network $\mathcal{S}$ using only few samples per category.
        For $K$-shot distillation, the optimization algorithm needs to search a large parameter space of student $\mathcal{S}$ with only $K$ samples per category. 
        Hence, it is hard to directly optimize the student network with plain knowledge distillation scheme.
        To accomplish this goal, we adopt network grafting strategy to decompose the student network into several blocks, each of which only contains fewer parameters to be optimized.

        Let $\mathcal{D}_c = \{x_{c,k}\}_{k=1}^{K}$ denotes the $K$ samples from class $c$. 
        The whole training dataset for $K$-shot $N$-way distillation can be presented as $\mathcal{D}=\cup_{c=1}^N \mathcal{D}_c$.
        The teacher network can be regarded as a composite function as $\mathcal{T}(x) = f_L \circ \cdots \circ f_l \circ \cdots \circ f_1(x)$, where $f_l$ denotes the $l$-th block of teacher.
        In the same way, the student network can be presented as $\mathcal{S}(x) = h_L \circ \cdots \circ h_l \circ \cdots \circ h_1(x)$, where $h_l$ denotes the $l$-th block of student.
        The knowledge distillation can be decomposed into a series of block distillation problems, where student block $h_l$ learns to master the functions of the corresponding block in the teacher: $f_l$. 

        To this end, we adopt a dual-stage knowledge distillation strategy, which is depicted in Figure \ref{fig:overview}. 
        In the first stage, each block of student is grafted onto teacher network, which replaces the corresponding block of teacher. 
        Note that only single block of the teacher is replaced with the corresponding student block each time. 
        Then the grafted student block is trained with the distillation procedure between the grafted teacher network and the original teacher one.
        In the second stage, all the trained student blocks are progressively grafted on teacher network, until the whole teacher network is fully replaced by a series of student block.

    }

    \subsection{Block Grafting} \label{sec:block_graft}
    {
        In the setting of few samples available, it's believed that the optimization of neural network is difficult, especially the network with massive parameters.
        To reduce the complexity of network optimization, the student network is decomposed into a series of blocks, each of which contains fewer parameters.
        For the training of block module, two alternative solutions are available, as shown in Figure~\ref{fig:block_grafting}.

        \begin{figure*}[t]
        \centering
        \includegraphics[width=0.95\linewidth]{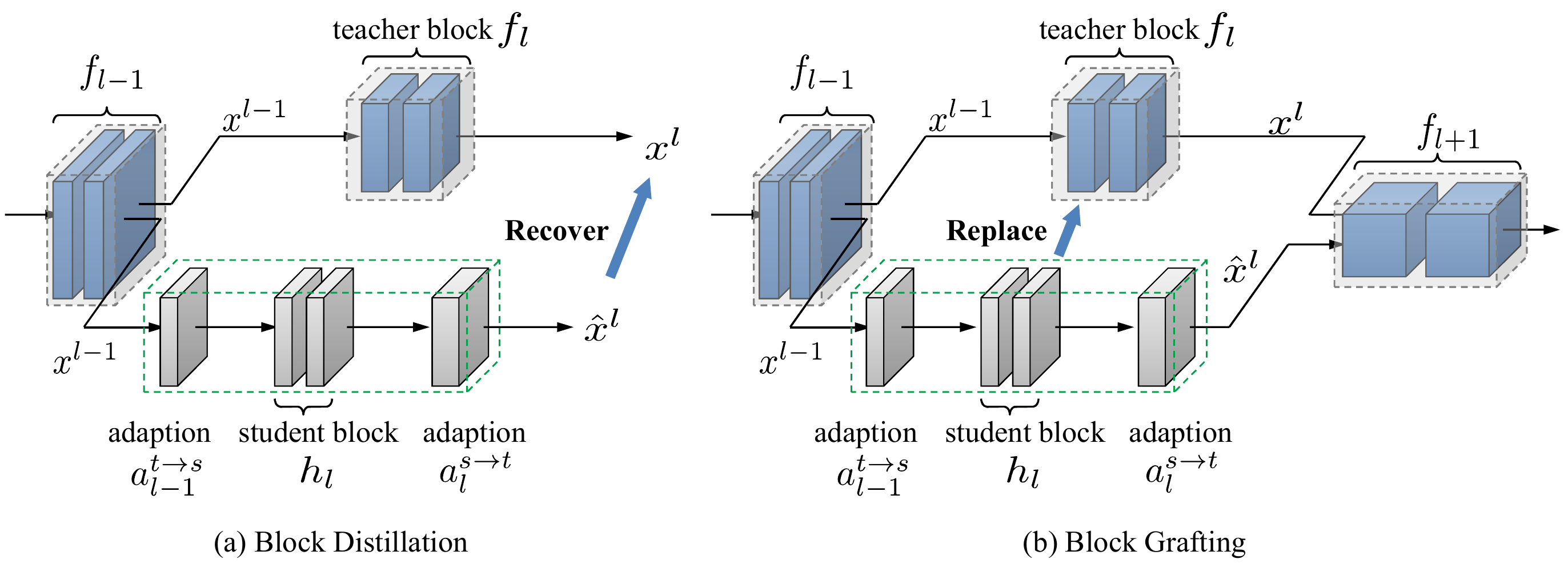}
        \caption{
        Two alternative block-wise distillation solutions. 
        (a) Block distillation tries to recover the output of teacher block using a lightweight student block.
        (b) Block grafting is trained to replace teacher block with student one to recover the final prediction of teacher.
        Note that only the modules in green dashed box are updated during optimization.
        }
        \label{fig:block_grafting}
        \end{figure*}

        For the first solution, the input and output of the corresponding block from teacher network are used to train the student block, as depicted in Figure~\ref{fig:block_grafting} (a). 
        This solution aims to imitate and recover the output of teacher block originally.
        However, due to model compression, the student block is generally much smaller than the teacher one.
        In other word, the capacity of student block is smaller than teacher block.
        There is some noise in teacher block output, which is trivial for final classification decision. 
        Learning without information selection may hurt the performance of the student. 

        For the second solution, the student block is grafted on teacher network, as shown in Figure~\ref{fig:block_grafting} (b).
        The grafted teacher network can be denoted as:
        \begin{equation} \label{eq:block_graft}
            \mathcal{T}_l^B(x) = f_L \circ \cdots f_{l+1} \circ h_l \circ f_{l-1} \circ \cdots f_1(x),
        \end{equation}
        where the $l$-th block of student: $h_l$ replaces the teacher block $f_l$.
        Note that the block number of student needs to be equal to the teacher one, but the inner structure of the $h_l$ and the $f_l$ can be different.
        The grafted teacher $\mathcal{T}_l^B$ is trained to recover the predictions of the original teacher $\mathcal{T}$ and only the parameters of $h_l$ are optimized.
        This solution encourages the block $h_l$ to focus on the knowledge, which is vital for final classification predictions.
        It effectively avoids the problem in the first solution, where the block only learns the original outputs of teacher block but not the function of the block in the whole network.

        Due to the channel dimension difference between student block and teacher one, we introduce adaption module to align the channel dimension difference.
        The adaption module can be categorized into two kinds: adaption module for the dimension transform from teacher block to student one $a_{l-1}^{t \rightarrow s}(x^{l-1})$ and the one from student to teacher $a_{l}^{s \rightarrow t}(x^{l})$.
        Combined with adaption modules, the wrapped scion $\mathcal{H}_l(x^{l-1})$ can be written as:
        \begin{equation} \label{eq:adaption}
            \mathcal{H}_l(x^{l-1}) = a_{l}^{s \rightarrow t}\circ h_l \circ a_{l-1}^{t \rightarrow s}(x^{l-1}), 
        \end{equation}
        where $x^l$ denotes the output of $l$-th block from network.
        The adaption module is implemented with $1 \times 1$ convolution operation.
        It achieves a linear recombination of input features across channel dimension and doesn't change the size of receptive field, which is designed to align features between student and teacher.
        There are two special cases to be clarified: 
        \begin{equation} \label{eq:special_case}
            \mathcal{H}_1(x) = a_{1}^{s \rightarrow t}\circ h_1(x), 
            \mathcal{H}_L(x) = h_L(x) \circ a_{L-1}^{t \rightarrow s}(x^{L-1}), 
        \end{equation}
        where $\mathcal{H}_1(x)$ and $\mathcal{H}_L(x)$ denote the first wrapped scion and the last one, respectively.
        Combining Eq.\ref{eq:block_graft}, Eq.\ref{eq:adaption} and Eq.\ref{eq:special_case}, the final grafted teacher can be written as:
        \begin{equation} \label{eq:final_block_graft}
            \mathcal{T}_l^B(x) = f_L \circ \cdots f_{l+1} \circ \mathcal{H}_l \circ f_{l-1} \circ \cdots f_1(x).
        \end{equation}

    }

    {
        \setlength{\textfloatsep}{2pt}
        \renewcommand{\algorithmicrequire}{\textbf{Input:}} 
        \renewcommand{\algorithmicensure}{\textbf{Output:}}
        \begin{algorithm}[!t]
            \caption{Network Grafting for Few-Shot Knowledge Distillation}
            \label{alg:A1}
            \begin{algorithmic}[1]
                \Require{Teacher model $\mathcal{T}$; 
                        Few unlabeled training data $\mathcal{D} = \{x_i\}_{i=1}^{N \cdot K}$.}
                \Ensure{The parameters of student block $\mathcal{S}$}.
                \For{$l=1$ to $L$}
                    \State Wrap $h_l$ in $a_l^{s \rightarrow t}$ and $a_{l-1}^{t \rightarrow s}$ to obtain $\mathcal{H}_l$ by Eq.~\ref{eq:adaption} and Eq.~\ref{eq:special_case};
                    \State Graft $\mathcal{H}_l$ onto $\mathcal{T}$ to obtain $\mathcal{T}_l^B$ by Eq.~\ref{eq:final_block_graft};
                    \For{number of training epochs}
                        \State Feed $x$ into $\mathcal{T}$ and $\mathcal{T}_l^B$ to compute $\mathcal{L}_l^{B}$ by Eq.~\ref{eq:loss_block_graft};
                        \State Update the parameters of $\mathcal{H}_l$ using Adam;
                    \EndFor
                \EndFor
                \State Initialize grafted teacher by $\mathcal{T}_{1}^N \leftarrow \mathcal{T}_{1}^B$;
                \For{$l=2$ to $L$} 
                    \State Combine $\mathcal{T}_{l-1}^N$ and $\mathcal{H}_l$ to obtain $\mathcal{T}_l^N$ by Eq.~\ref{eq:net_graft};
                    \For{number of training epochs}
                        \State Feed $x$ into $\mathcal{T}$ and $\mathcal{T}_l^N$ to compute $\mathcal{L}_l^{N}$ by Eq.~\ref{eq:loss_net_graft};
                        \State Update the parameters of $\{ \mathcal{H}_j \}_{j=1}^l$ using Adam;
                    \EndFor
                \EndFor

                \State Merge $\{a_l^{t \rightarrow s}\}_l$ and $\{a_l^{s \rightarrow t}\}_l$ into $\{h_l\}_l$ and obtain $\mathcal{T}_L^{N}$, namely $\mathcal{S}$, by Eq.~\ref{eq:final_net_graft}.
            \end{algorithmic}
        \end{algorithm}
    }

    \subsection{Progressive Network Grafting}
    {
        A series of trained student blocks can be obtained from the above section, each of which can make decision with the blocks of teacher network.
        However, these trained blocks are not trained to work with each other.
        In this section, we adopt a network grafting strategy to progressively increase the student blocks in the grafted teacher network and reduce the dependency of the original teacher. 

        On the base of $\mathcal{T}_1^B(x) = f_L \circ \cdots \circ f_2 \circ \mathcal{H}_1(x)$, the trained blocks: $\mathcal{H}_2, \mathcal{H}_3, \cdots, \mathcal{H}_L$ are sequentially grafted onto $\mathcal{T}$, as shown in Figure~\ref{fig:overview}. 
        The grafted teacher for network grafting can be denoted as:
        \begin{equation} \label{eq:net_graft}
            \mathcal{T}_l^N(x) = f_L \circ \cdots f_{l+1} \circ \mathcal{H}_l \circ \mathcal{H}_{l-1} \circ \cdots \mathcal{H}_1(x),
        \end{equation}
        where $\mathcal{T}_1^N(x) = \mathcal{T}_1^B(x)$.
        During network grafting, a sequence of models: $\{\mathcal{T}_l^N(x)\}_{l=1}^L$ are optimized.
        Finally, $\mathcal{T}_L^N(x)$ is obtained, which connects all student blocks and forms a complete network.
        However, $\mathcal{T}_L^N(x)$ is still different from original student network $\mathcal{S}(x)$.
        $\mathcal{T}_L^N(x)$ is composed of a series of $\mathcal{H}_l$, but $\mathcal{S}(x)$ is composed of $h_l$.
        Compared to $h_l$, $\mathcal{H}_l$ contains additional adaption module: $a_{l}^{s \rightarrow t}$ or $a_{l-1}^{t \rightarrow s}$, which means the obtained $\mathcal{T}_L^N(x)$ has more parameters than $\mathcal{S}(x)$.

        Thanks to the linearity of adaption module, the parameters of adaption module can be merged into the convolution layer in the next block $h_{l+1}$ without increasing any parameters. 
        For $\mathcal{H}_{l+1} \circ \mathcal{H}_l = a_{l+1}^{s \rightarrow t}\circ h_{l+1} \circ a_{l}^{t \rightarrow s} \circ a_{l}^{s \rightarrow t}\circ h_l \circ a_{l-1}^{t \rightarrow s}$, $(a_{l}^{t \rightarrow s} \circ a_{l}^{s \rightarrow t})$ can be merged into $h_{l+1}$. We denote $\hat{h}_{l+1} = h_{l+1} \circ a_{l}^{t \rightarrow s} \circ a_{l}^{s \rightarrow t}$.
        Then, $\mathcal{T}_L^N(x)$ can be transformed into the following form:
        \begin{equation} \label{eq:final_net_graft}
            \mathcal{T}_L^N(x) = \hat{h}_L \circ \cdots \hat{h}_{l+1} \circ \hat{h}_l \circ \hat{h}_{l-1} \circ \cdots \hat{h}_1(x),
        \end{equation}
        which has the same network architecture as $\mathcal{S}$. 
        In other word, we achieve the knowledge transfer from teacher $\mathcal{T}$ to student $\mathcal{S}$.

    }

{
    \begin{table*}[ht]
    \centering
    \begin{threeparttable}
    {
        \begin{tabular}{c|c|ccc|ccc}
        \hline
        \multicolumn{1}{c|}{} & \multicolumn{1}{c|}{}    & \multicolumn{3}{c|}{\textbf{CIFAR10}}  & \multicolumn{3}{c}{\textbf{CIFAR100}}\\
        \textbf{Method} & \textbf{\# Samples} & \textbf{\# Params} & \textbf{\# FLOPs} & \textbf{Accuracy} (\%) & \textbf{\# Params} & \textbf{\# FLOPs} & \textbf{Accuracy} (\%)\\ 
        \hline
        Teacher (VGG16) 
                & 5,000 & 15.0M   & 0.31G   & 92.83                   & 15.0M   & 0.31G   & 69.82 \\
        KD (full)
                & 5,000 & 5.4M    & 0.22G   & 92.06$\pm$0.17          & 5.4M    & 0.22G   & 68.31$\pm$0.15\\
        FitNet (full)
                & 5,000 & 5.4M    & 0.22G   & 92.75$\pm$0.11          & 5.4M    & 0.22G   & 70.12$\pm$0.12\\
        \hline
                & 1     &         &         & 71.80$\pm$1.84          &         &         & 37.73$\pm$1.53 \\
        KD (few)
                & 5     & 5.4M    & 0.22G   & 87.49$\pm$1.01          & 5.4M    & 0.22G   & 64.03$\pm$0.06 \\
                & 10    &         &         & 88.48$\pm$0.16          &         &         & 65.27$\pm$0.17 \\
        \hline
                & 1     &         &         & 74.43$\pm$1.43          &         &         & 41.67$\pm$1.23 \\
        FitNet (few)
                & 5     & 5.4M    & 0.22G   & 88.53$\pm$0.92          & 5.4M    & 0.22G   & 64.13$\pm$0.34 \\
                & 10    &         &         & 88.76$\pm$0.14          &         &         & 65.64$\pm$0.15 \\
        \hline
                & 1     &         &         & 87.42$\pm$0.84          &         &         & 48.57$\pm$1.14 \\
        FSKD
                & 5     & 5.4M    & 0.22G   & 88.83$\pm$0.63          & 5.4M    & 0.22G   & 65.47$\pm$0.18 \\
                & 10    &         &         & 92.37$\pm$0.00$\dag$    &         &         & 67.34$\pm$0.12 \\
        \hline
                & 1     &         &         & 69.57$\pm$1.39          &         &         & 41.32$\pm$0.18 \\
        Cross Distillation
                & 5     & 5.4M    & 0.22G   & 84.91$\pm$0.98          & 5.4M    & 0.22G   & 63.81$\pm$0.15 \\
                & 10    &         &         & 86.61$\pm$0.71          &         &         & 64.95$\pm$0.24 \\
        \hline
                & 1     &         &         & \textbf{90.74$\pm$0.49}          &         &         & \textbf{64.22$\pm$0.17} \\
        Ours (VGG16-half)
                & 5     & 5.4M    & 0.22G   & \textbf{92.88$\pm$0.07}          & 5.4M    & 0.22G   & \textbf{68.16$\pm$0.20} \\
                & 10    &         &         & \textbf{92.89$\pm$0.06} &         &         & \textbf{68.86$\pm$0.03} \\
        \hline
                & 1     &         &         & 73.69$\pm$0.91          &         &         & 55.51$\pm$0.92 \\
        Ours (ResNet18)
                & 5     & 11.2M   & 0.22G   & 91.79$\pm$0.34          & 11.2M   & 0.22G   & 66.64$\pm$0.12 \\
                & 10    &         &         & 92.48$\pm$0.13          &         &         & 67.77$\pm$0.04 \\
        \hline
        \end{tabular}
        }
    \begin{tablenotes}
    \item[$\dag$] denotes the experimental result reported in~\cite{li2018few}.
    \end{tablenotes}
    
    \caption{The performance of few-shot distillation on CIFAR dataset.
            ``full'' denotes training with full dataset. ``few'' denotes few-shot training.
            The second column: ``\# Samples'' denotes the number of samples per class. All student networks adopt VGG16-half, except ``Ours (ResNet18)''.
            All results are averaged over five runs and error bars correspond to the standard deviation.}
    \label{table:cifar}
    \end{threeparttable}
    \end{table*}
}

    \subsection{Optimization}
    {

        The scale of logits from different network architectures may have a large gap, which may result in optimization difficulty. 
        To this end, we propose an $\rm{L}_2$ loss function on normalized logits for knowledge transfer between teacher block $f_l$ and student block $h_l$ as follow:
        \begin{equation} \label{eq:loss_block_graft}
            \mathcal{L}_l^{B}(x) = \frac{1}{N} \Vert \widetilde{\mathcal{T}}_l^B(x) - \widetilde{\mathcal{T}}(x) \Vert_2^2,
        \end{equation}
        where $\widetilde{\mathcal{T}}(x) = \left.{\mathcal{T}(x)}\middle/{\Vert \mathcal{T}(x) \Vert_2}\right.$.
        In the optimization of block grafting, only the wrapped student block $\mathcal{H}_l$ is learnable and updated with the gradient $\nabla_{\Theta_l} \mathcal{L}_l^{B}$, where $\Theta_l$ denotes the parameters of $\mathcal{H}_l$.
        For network grafting, a similar distillation is adopted,
        \begin{equation} \label{eq:loss_net_graft}
            \mathcal{L}_l^{N}(x) = \frac{1}{N} \Vert \widetilde{\mathcal{T}}_l^N(x) - \widetilde{\mathcal{T}}(x) \Vert_2^2.
        \end{equation}
        The difference to block grafting is that a sequence of wrapped student blocks: $\{ \mathcal{H}_i \}_{i=1}^l$ need to be optimized, for the $l$-th step of network grafting.
        The complete algorithm for our proposed approach can be summarized as Algorithm~\ref{alg:A1}.
        
    }
}

\section{Experiments}
{
    \subsection{Experimental Settings}
    {
        \subsubsection{Datasets and Models.}
        {
            Both CIFAR10~\cite{krizhevsky2009learning} and CIFAR100~\cite{krizhevsky2009learning} are composed of $60,000$ colour images with $32 \times 32$ size, where $50,000$ images are used as training set and the rest $10,000$ images are used as test set.
            The CIFAR 10 dataset contains 10 classes, and the CIFAR100 contains 100 classes.
            In few-shot setting, we randomly sample $K$ samples per class from the original CIFAR datasets as training set, where $K \in \{1, 5, 10\}$.
            Random crop and random horizontal flip are applied to training images to augment the dataset.
            The test set is the same as the original one.
            Due to low resolution of images in CIFAR, a modified VGG16~\cite{li2016pruning} are used as the teacher model.
            Following \cite{li2018few} and \cite{li2016pruning}, we adopt VGG16-half as student model.

            The ILSVRC-2012 dataset~\cite{ILSVRC15} contains 1.2 million images as training set, $50,000$ images as validation set, from $1,000$ categories. 
            We randomly sample 10 images per class from the training set for few-shot training.
            The training images are augmented with random crop and random horizontal flip.
            During test, the images are cropped into $224 \times 224$ size in the center.
            We adopt PyTorch official trained ResNet34~\cite{he2016deep} 
            as the teacher model, ResNet18 as the student one.

        }

        \subsubsection{Implementation Details.}
        {
            The proposed method is implemented using PyTorch on a Quadro P6000 24G GPU.
            The batch size is set to 64 for 10-shot training. 
            For $K$-shot training, we set batch size to $\lfloor 64\times K/10 \rfloor$.
            For all experiments, we adopt Adam algorithm for network optimization. 
            Without extra clarification, the following learning rates work when batch size is 64. 
            For other batch size: $B$, the learning rate is scaled by the factor: $B/64$ as~\cite{he2016deep}.
            The learning rates for block grafting and network grafting on CIFAR10 are set to $2.5 \times 10^{-4}$, $1 \times 10^{-4}$, respectively. 
            For CIFAR100, the learning rates are $1 \times 10^{-3}$, $5 \times 10^{-5}$, respectively. 
            Following \cite{kingma2014adam}, we set the weight decay to zero and the running averages of gradient and its square to 0.9 and 0.999, respectively.
            We adopt the weight initialization proposed by \cite{he2015delving}.
            For ResNet18 on ILSVRC-2012, we adopt the same optimizer and weight initialization method as VGG16-half.
            During block grafting, we set the learning rate of \emph{block1} and \emph{block2} to $10^{-4}$, the one of \emph{block3} and \emph{block4} to $10^{-3}$.
            During network grafting, the learning rates for \emph{block1$\sim$2}, \emph{block1$\sim$3} and \emph{block1$\sim$4} are set to $10^{-4}$, $2 \times 10^{-3}$ and $10^{-3}$, respectively.
        }

    }

    \subsection{Experimental Results}
    {
        {
        \subsubsection{Homogeneous Architecture Knowledge Distillation.}
        Intuitively, the knowledge between homogeneous network architectures tends to have a more clear correlation than heterogeneous networks, especially for block-wise distillation situation. 
        In this section, we first investigate knowledge distillation between homogeneous network to validate the effectiveness of the proposed method.
        In Table~\ref{table:cifar}, VGG16-half has a similar network architecture as VGG16 but has fewer channels than VGG16 in the corresponding layer.
        With full dataset, both KD~\cite{hinton2015distilling} and FitNet~\cite{romero2015fitnets} for VGG16-half achieve comparable performance as their larger teacher.
        However, when only few samples are available, both KD and FitNet encounter a significant performance drop.
        FSKD~\cite{li2018few} adopts a dual-step strategy: first network pruning and then block-wise distillation, which achieves an efficient improvement against KD and FitNet in few-shot setting.
        Our proposed method achieves comparable performance as KD trained with full dataset, even better than teacher in 10-shot setting on CIFAR10.
        And our method is also superior to other few-shot knowledge distillation methods in all settings.
        FSKD focuses on the imitation of the intermediate teacher outputs, whose student is not optimized to imitate the predictions of teacher.
        We owe the improvement of our method to the end-to-end imitation of the teacher predictions, which is more robust to some trivial intermediate noise.

        \subsubsection{Heterogeneous Architecture Knowledge Distillation.}
        To further verify our method, we conduct few-shot distillation between heterogeneous networks. 
        We adopt ResNet18 as the student, whose network architecture is significantly different from VGG16.
        Despite more parameters than VGG16-half, the ResNet18 does not achieve better performance in few-shot distillation setting than VGG16-half. 
        Especially when only one sample is available per category, the performance drop of ResNet18 is dramatic on both CIFAR10 and CIFAR100. 
        We guess that the knowledge distribution of ResNet18 is significantly different from the VGG16 one across blocks, which harms the performance of knowledge distillation. 
        We put the exploration for knowledge transfer between very different network architectures in future work.

        \subsubsection{Knowledge Distillation for Large-Scale Dataset.}
        To validate the generality of the proposed method, we also conduct experiments on a more challenging large-scale dateset: ILSVRC-2012. 
        As shown in Table~\ref{table:imagenet}, the experimental results demonstrate that our proposed method significantly outperforms other few-shot distillation baselines. 
        The performance of our method is slightly below ResNet18 trained on full ILSVRC-2012 with classification loss. 
        We deduce that the dataset for few-shot distillation struggles to cover the high diversity of full ILSVRC-2012.
        
        \begin{table}[ht]
        \centering
        \resizebox{\linewidth}{!}
        {
            \begin{tabular}{llcccccc}
            \toprule
            \textbf{Method} & \textbf{\#Samples} & \textbf{\#Params} & \textbf{\#FLOPs} & \textbf{Acc@1} & \textbf{Acc@5}\\ 
            \hline
            Teacher (ResNet34)  
                                & 1,000           & 21.8M         & 3.6G     & 73.31\%        &  91.42\%    \\
            ResNet18 (xenc)   
                                & 1,000           & 11.7M         & 1.8G     & 69.76\%        &  89.08\%    \\
            \hline
            KD (ResNet18)
                                & 10              & 11.7M         & 1.8G     & 46.40\%        &  73.71\%   \\
            FitNet (ResNet18)  
                                & 10              & 11.7M         & 1.8G     & 45.82\%        &  73.46\%   \\
            Ours (ResNet18)
                                & 10              & 11.7M         & 1.8G     & 68.15\%        &  88.83\%   \\
            \bottomrule
            \end{tabular}   

        }
        \caption{The performance of few-shot distillation on ILSVRC-2012 dataset. 
        ``Acc@1'' and ``Acc@5'' denote top 1 accuracy and top 5 accuracy, respectively.
        ``xenc'' denotes training with cross entropy classification loss.}
        \label{table:imagenet}
        \end{table}
        }

        \begin{figure*}[ht]
        \centering
        \subfloat[1-Shot Distillation]{{\includegraphics[width=0.33\linewidth]{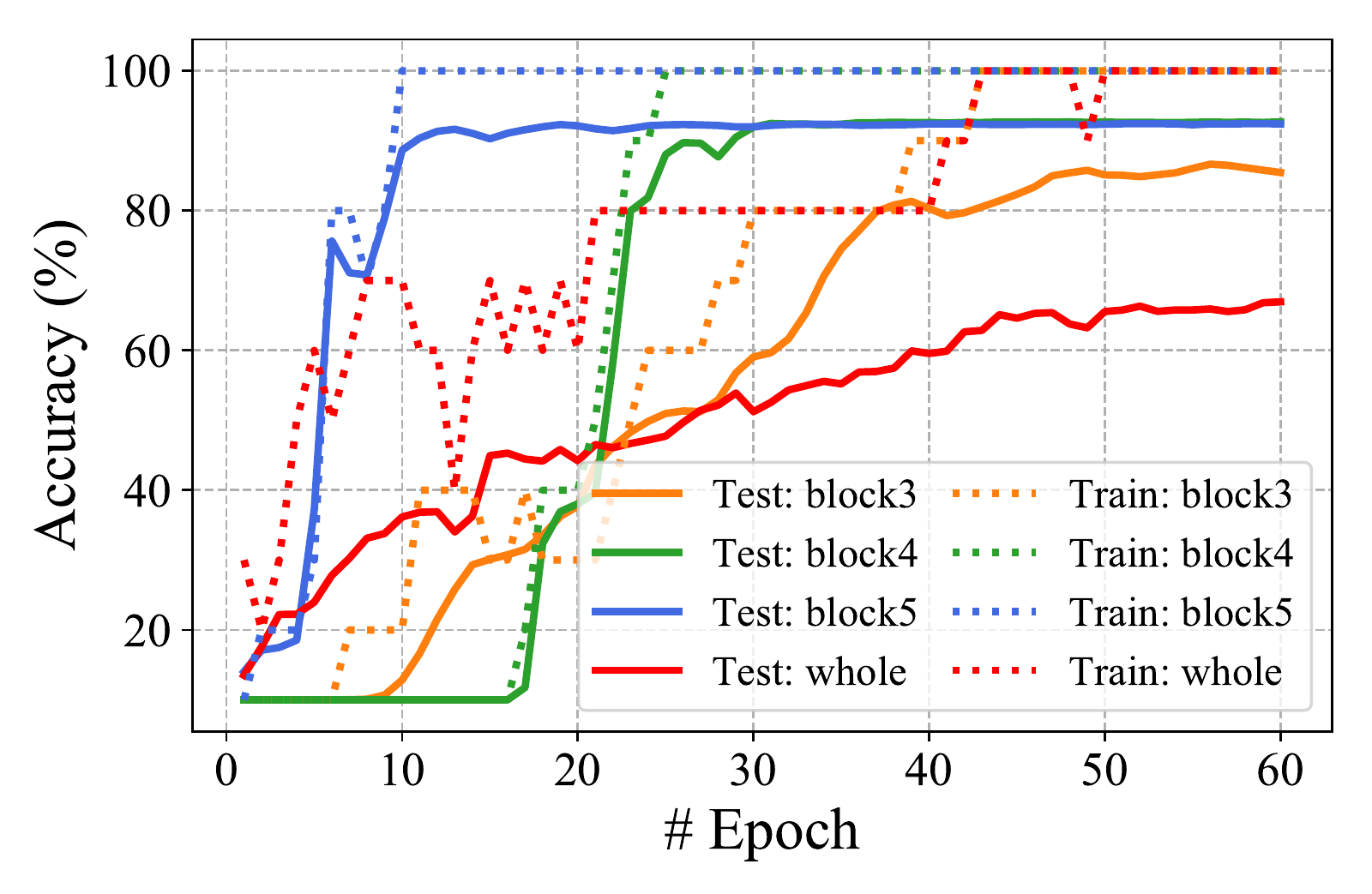}}}
        \subfloat[5-Shot Distillation]{{\includegraphics[width=0.33\linewidth]{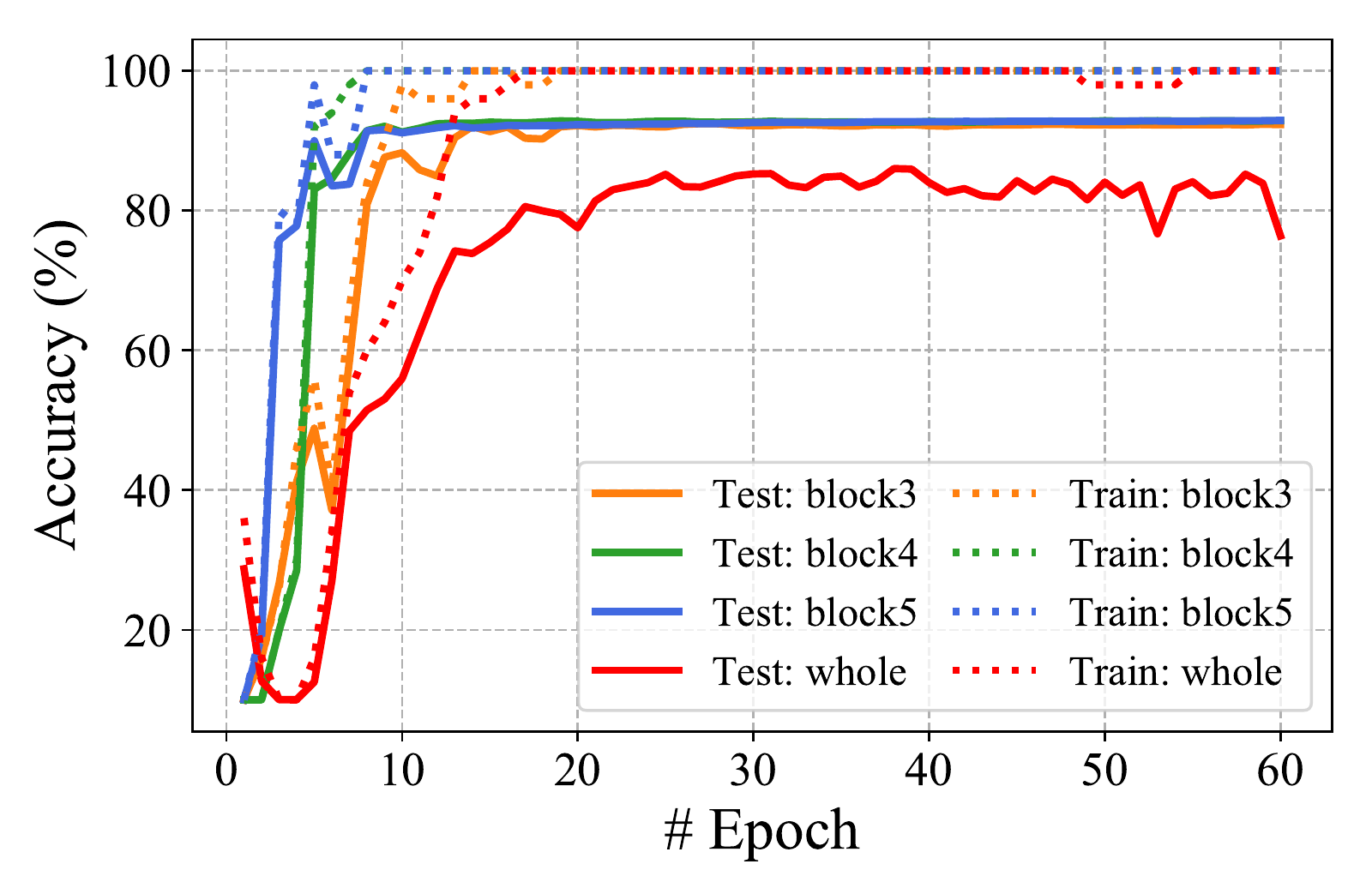}}}
        \subfloat[10-Shot Distillation]{{\includegraphics[width=0.33\linewidth]{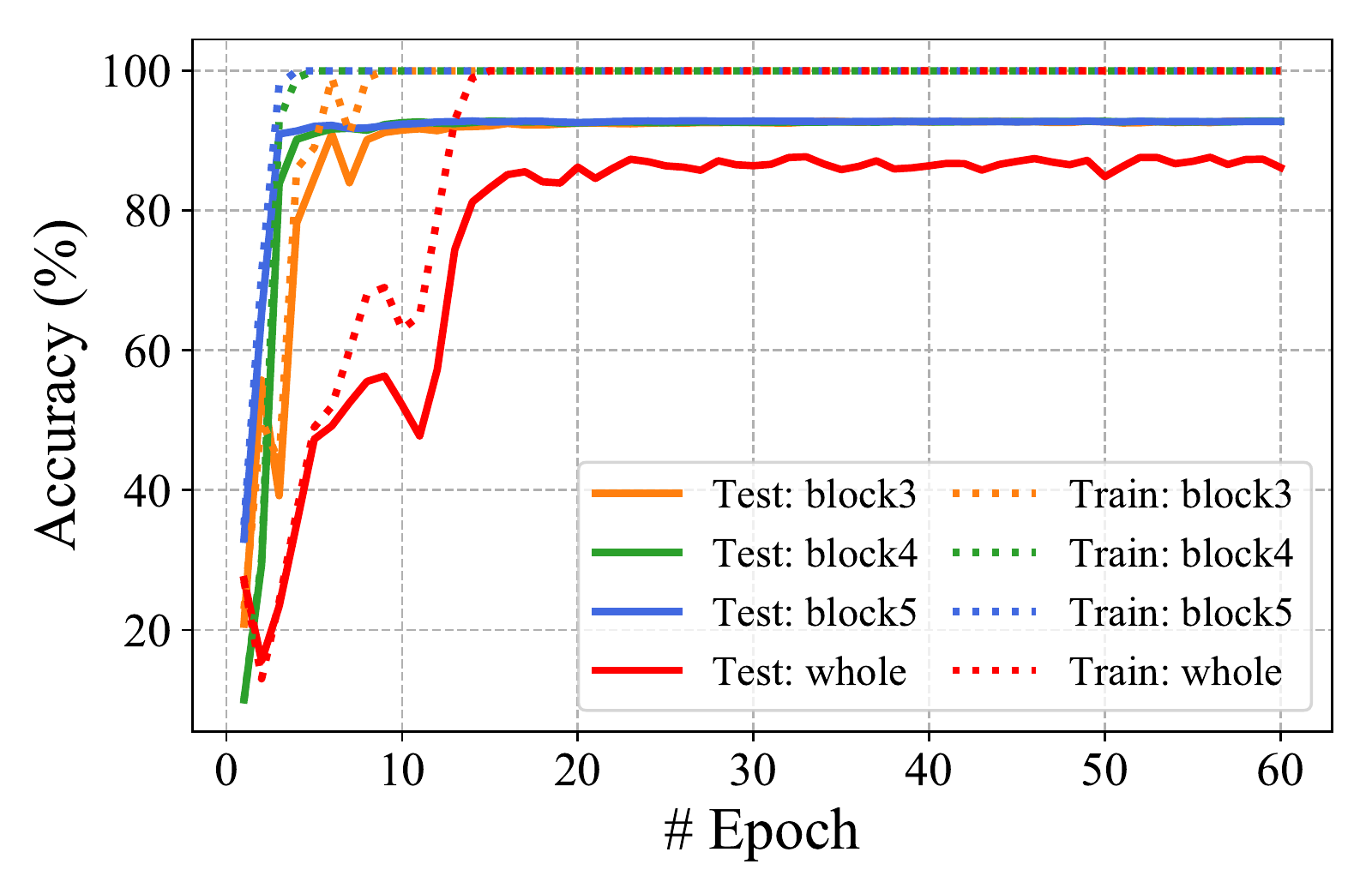}}}
        \caption{
        The comparison of optimization efficiency on block grafting and the whole student on CIFAR10.
        }
        \label{fig:efficiengy_block_grafting}
        \end{figure*}

        \subsubsection{Learning with Different Numbers of Samples.}
        {
            To investigate the effect of different numbers of training samples, we conduct a series of few-shot distillation experiments. 
            The experiments are implemented on CIFAR10 and CIFAR100, as shown in Figure~\ref{fig:num_sample}.
            For KD and FitNet, the distillation performances are significantly improved with the increment of the training sample number.
            When the number of training samples is extremely small, the performances of KD and FitNet are both far from the teacher one.
            Our proposed method achieves comparable performance as teacher, even when only one training sample is available.
            Due to smaller parameter search space, our method is more robust to the size of training dataset.

            \begin{figure}[ht]
            \centering
            \subfloat[CIFAR10]{{\includegraphics[width=\linewidth]{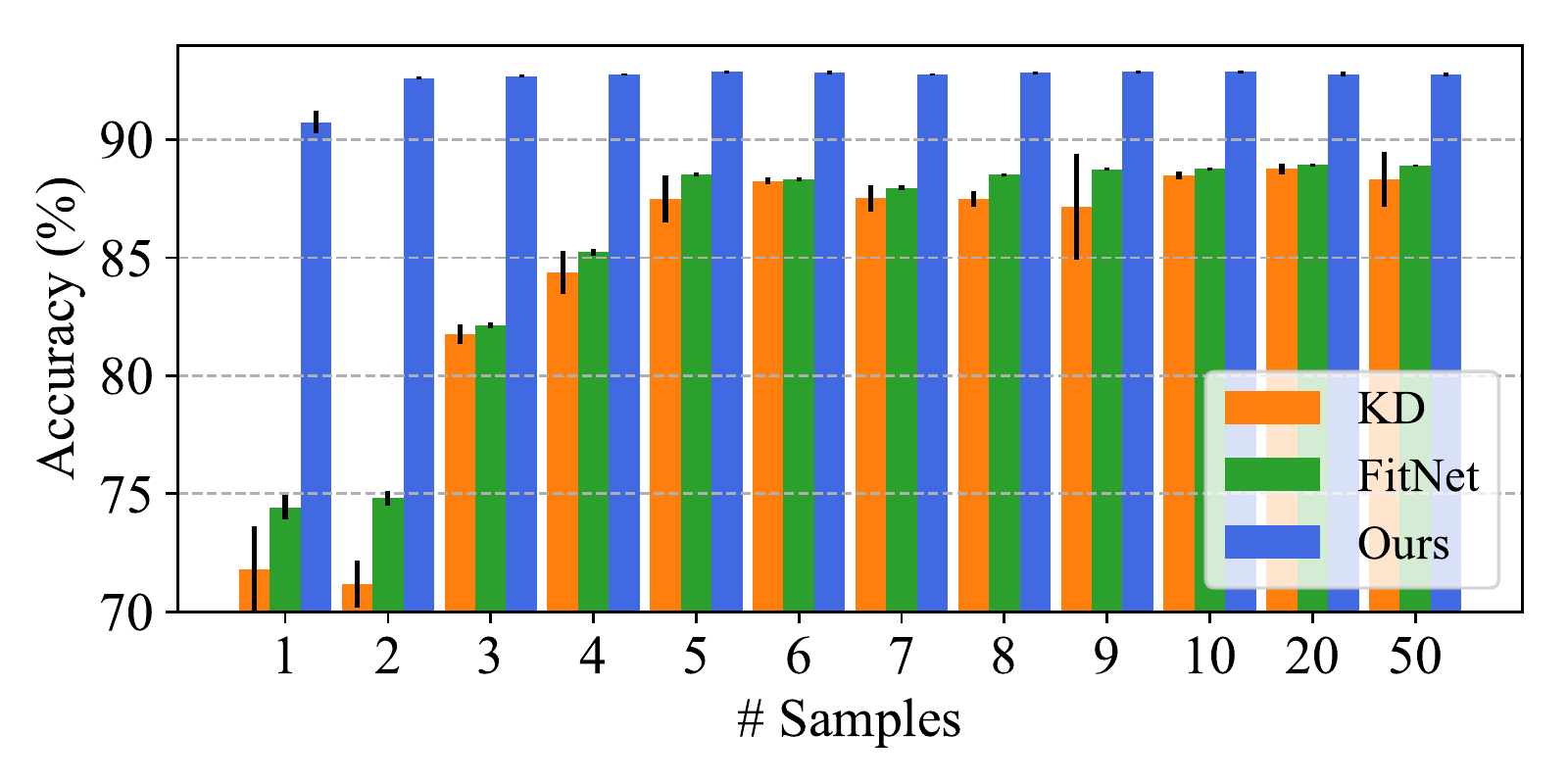}}} \\
            \subfloat[CIFAR100]{{\includegraphics[width=\linewidth]{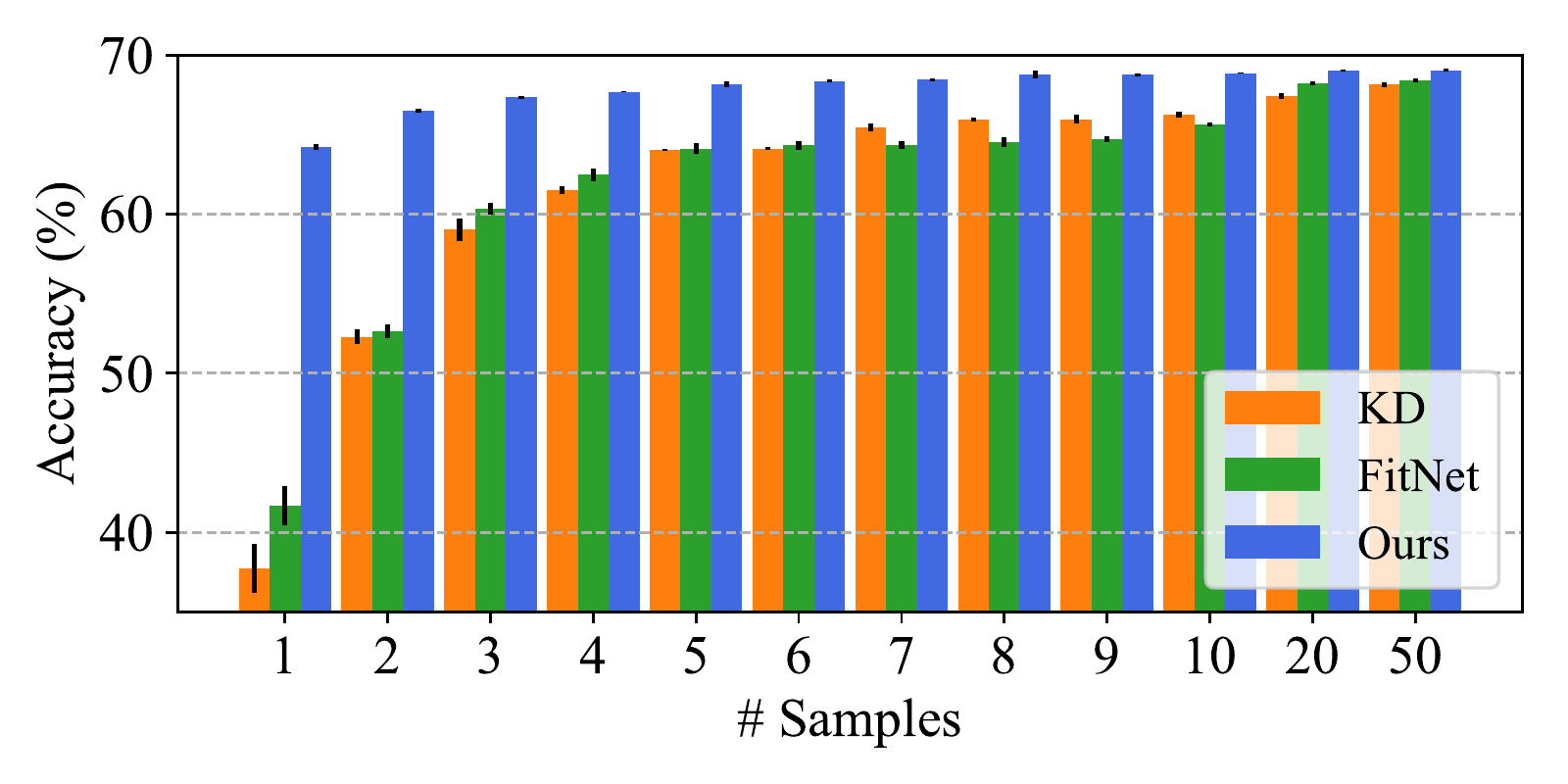}}}
            \caption{
            Learning with different numbers of samples.
            All experiments are averaged over five runs and error bars correspond to the standard deviation.
            }
            \label{fig:num_sample}
            \end{figure}

        }

        \subsubsection{Optimization Efficiency of Block Grafting.}
        {
            To evaluate the efficiency of our proposed block grafting strategy, we compare the training curves of single grafted block optimization and the whole student optimization, denoted as ``whole'', in different few-shot settings.
            Specifically, the last three blocks of VGG16-half, which contain most of parameters in network, are selected as the target to validate block grafting algorithm. 
            As shown in Figure~\ref{fig:efficiengy_block_grafting}, our proposed method not only converges more quickly, but also outperforms the ``whole'' by a large margin. 
            We observe that our method has a smaller gap between training accuracy and test one, which demonstrates our method can significantly reduce the risk of overfitting. 

        }

        \subsubsection{Partial Network Grafting.}
        {
            Some trained networks contain several cumbersome blocks, which can be simplified. 
            Partial network grafting provides an alternative solution to improve the compactness of the existing trained network. 
            We graft some lightweight blocks onto the target trained network to reduce the original model size.
            In this section, we evaluate the effect of different blocks' graft from ResNet18 to ResNet34, as shown in Table~\ref{table:partial}.
            We split the backbone of ResNet34 and ResNet18 into four blocks: block1$\sim$4, according to down-sampling layer.
            As depicted in Table~\ref{table:partial}, the \emph{block3} and \emph{block4} contain most of parameters in ResNet34.
            Replacing these blocks with the corresponding blocks from ResNet18 can significantly reduce the original model size.
            For example, replacing \emph{block3} with the one from ResNet18 can reduce the parameter number from 6.82M to 2.10M, a 69.2\% reduction.
            Compared to the ResNet18 trained on full ILSVRC-2012, the performance only drops 1.65\%, where only 10 samples per category are used to train network.

            \begin{table}[h]
            \centering
            \resizebox{\linewidth}{!}
            {
            \begin{tabular}{lccccccc}
            \hline
            \toprule
            & \textbf{original}    & \textbf{block1}    & \textbf{block2}   & \textbf{block3}  & \textbf{block4} \\ 
            \hline
            \textbf{Params} (M) &  / & 0.22$\rightarrow$0.14 & 1.12$\rightarrow$0.52 & 6.82$\rightarrow$2.10 & 13.11$\rightarrow$8.40\\
            &  / & 36.4\%$\downarrow$ & 53.6\%$\downarrow$ & 69.2\%$\downarrow$ & 35.9\%$\downarrow$ \\
            \hline
            \textbf{Accuracy} (\%) & 73.31             & 72.55              &  70.33            &  68.11          & 68.40  \\
            \bottomrule
            \end{tabular}
            }

            \caption{The performance of block grafting from ResNet18 to ResNet34 on ILSVRC-2012. 
            ``original'' denotes the original teacher network.
                    }
            \label{table:partial}
            \end{table}
        }
    }

}

\section{Conclusion and Future Work}
{
    In this paper, we propose an progressive network grafting method to distill knowledge from teacher with few unlabeled samples per class. 
    This is achieved by a dual-stage approach.
    In the first stage, the student is split into blocks and grafted onto the corresponding position of teacher.
    The student blocks are optimized with fixed teacher blocks using distillation loss.
    In the second stage, the trained student blocks are incrementally grafted onto the teacher and trained to connect to each other, until the whole student network replaces the teacher one.
    Experimental results on several benchmarks demonstrate that the proposed method successfully transfers knowledge from teacher in few-shot setting and achieves comparable performance as knowledge distillation using full dataset. 

    Few-shot distillation can significantly reduce the training cost of neural network.
    For future work, we plan to explore network architecture search on distillation using block grafting, which aims to find a series of more efficient block modules.
    We believe that this work is one step toward efficient network architecture search.

}

\subsubsection{Acknowledgements.}
This work is supported by National Natural Science Foundation of China (U20B2066, 61976186),  the Major Scientifc Research Project of Zhejiang Lab (No. 2019KD0AC01) and Alibaba-Zhejiang University Joint Research Institute of Frontier Technologies.

\bibliographystyle{aaai}
\bibliography{mybib}

\end{document}